\documentclass[mlmain,onecolumn]{jmlr}

\usepackage{booktabs}

\jmlrproceedings{}{}   
\jmlrvolume{}
\firstpageno{1}
\jmlryear{2026}

\title[A Hub of Short Rows Inflates Intrinsic Dimension Estimation]{A Hub of
Short Rows Inflates Intrinsic Dimension Estimation of Token Embeddings}

\author{\Name{Alexandre Quemy} \Email{alexandre@hother.io}\\
\addr Hother Labs}

\begin{document}
\maketitle

\begin{abstract}

A token-embedding table holds a hub of short rows near its origin, and we show that
this cluster biases what nearest-neighbor intrinsic-dimension (ID) estimators report. 
Because of the concentration of measure, a token is closer to the central
cluster than to any other token, so its first two neighbors are both hub rows at nearly
the same distance. As a result, the ID estimators such as TwoNN return a dimension far above the real ID.
Measured one token at a time, dimension is a heavy-tailed distribution. Measured on the
full vocabulary, it grows with the model’s parameter count. However, when we remove the hub, the
heavy tail disappears and the measured dimension collapses to a narrow range for eleven
models, from GPT-2 to models such as K3 and GLM-4.7. The hub acts as a switch: a few hundred rows are 
enough to fully inflate the estimate.
We reproduced an experiment stating that the intrinsic dimension (ID) of Pythia's token-embedding
table grows with the parameter count, from $27$ to $122$ between 160M and 12B parameters. We show that
this result disappears when the hub is removed: the table then reads $10$ to $17$ at every size.
The hub contains a subset of the population that under-trained-token detectors flag, but
on Pythia the hub that we detected and removed as a whole was updated during training:
what seem to characterize these rows is simply their length, not an absence of updates. Finally, we show that
normalizing the rows instead of removing them gives the same lower reading.

\end{abstract}

\section{Introduction}
The intrinsic dimension of a token-embedding table depends on how it is
measured. One token at a time, it is heavy-tailed, with a median near
$389$ on GPT-2, which \citet{robinson2025manifold} interpret as a
violation of the manifold hypothesis. Taken on the whole vocabulary, it
is more modest but grows with the parameter count, from $25$ on Pythia-410M to $122$
on Pythia-12B~\citep{kataiwa2025tokenid}.
These numbers are used for instance to
assess how redundant a representation is~\citep{kataiwa2025tokenid}, how
large a subspace is worth training~\citep{aghajanyan2021intrinsic}, and
how models compare~\citep{valeriani2023geometry,razzhigaev2024shape}.

In this paper, we identify a population of tokens that connects this local and global
readings. When this hub of tokens is removed, the heavy tail disappears and the intrinsic
dimension becomes flat with the parameter count.


This population are rows with short norms, near the origin of
the table. We call it a hub, the term of \citet{radovanovic2010hubs} for
the few points that enter the nearest-neighbor lists of many others.
Across eleven models, from GPT-2 to K3 and GLM-4.7, we show that removing this set of tokens lowers the full-vocabulary estimate by up to $90\%$,
while control trims of the same size raise it. Conversely, adding back a few of the
 removed rows restores the whole effect. 
 In addition, the tail of the ID distribution observed in \citet{robinson2025manifold} mechanically disappears when the hub is removed.
 
On Pythia, we reproduce the published values of~\citet{kataiwa2025tokenid}, ID of $27$ to $122$ from 160M
to 12B, and the trim brings the ID estimate between $10$ and $17$. What
the estimator read was the depth of the norm tail, which deepens with parameter count. 

Finally, we found that normalizing every row to unit length agrees with the hub removal on nine tables. 
The argument is geometric: the normalization means using a 
cosine distance instead of the Euclidean distance for the ID estimators, which removes 
the hub's effect on the neighbor ratios.



\section{Related work}
\label{sec:related}
\textbf{Intrinsic-dimension estimators.} Nearest-neighbor
estimators infer a local dimension from near-neighbor distance ratios:
TwoNN from the ratio of second- to first-neighbor
distances~\citep{facco2017twonn}, the Levina--Bickel estimator from a full
$k$-neighbor set~\citep{levina2004mle}, later
corrected~\citep{mackay2005correction} and generalized to higher-order
ratios~\citep{denti2022gride}. Both assume a locally uniform sample from
a single manifold and both are exposed to hubness, the tendency of a few
points, near the data centroid, to enter the nearest-neighbor
lists of many others as the dimension grows~\citep{radovanovic2010hubs}.
It was first discovered in word-embedding spaces in cross-lingual
retrieval~\citep{dinu2015improving}. A correction is to rescale
distances~\citep{schnitzer2012mutual,feldbauer2019hubness}.
 Most of the literature use these estimators on hidden states, to study
 the profile across layers~\citep{ansuini2019intrinsic,valeriani2023geometry} but in this paper, we use it to measure the embedding table's intrinsic dimension.

\textbf{Dimension as a local, per-token quantity.} Local intrinsic
dimensionality~\citep{houle2017lid,amsaleg2015lid} and
Hidalgo~\citep{allegra2020hidalgo} attach a dimension to each point or
region. Closest to our work,
\citet{robinson2025manifold} estimate a dimension at every token of
GPT-2's table from how the number of neighbors grows with the radius. They
report a per-token dimension at small radius that is
heterogeneous and heavy-tailed (median $389$, interquartile $2$ to $531$). We discuss this result in Section~\ref{sec:defense}.

\textbf{Token-table geometry.} Token embeddings are known to be anisotropic, i.e., a few directions carry most of the variance between
tokens~\citep{mu2018allbutthetop,ethayarajh2019contextual,timkey2021bark,rudman2022isoscore}.
The same concentration persists in the text embeddings of current models,
which collapse toward an average token. \citet{wu2026unembedding}
trace it to the edges of the unembedding matrix's spectrum, where the most
frequent tokens are located.
The intrinsic dimension of a table has been measured with a Levina--Bickel estimator and found to grow from $25$ on
Pythia-410M to $122$ on Pythia-12B~\citep{kataiwa2025tokenid}. We repeat
that protocol before and after removing the hub.
Many of these tokens are already known to carry a distinct geometry of short
norms~\citep{land2024magikarp,rumbelow2023magikarp}, itself a face of the
frequency-driven degeneration of the embedding
space~\citep{gao2019degeneration}, though, as far as we know, their link to the table's
measured dimension has not been made before.

\section{Method and sources of dimension inflation}
\label{sec:instruments}
Every geometric quantity in this paper is computed from the input
embedding matrix $E \in \mathbb{R}^{V \times D}$. The unembedding matrix is left
for future work. 
This section defines how we measure the dimension of a
table and which rows the trim removes.

\textbf{The estimators.} TwoNN~\citep{facco2017twonn} forms, for each
point, the ratio $\mu = r_2 / r_1$ of its second- to first-neighbor
distances. On a locally uniform sample from a single manifold of dimension
$d$ the ratios follow a Pareto law. Therefore, the maximum-likelihood estimation is
$\hat d = N / \sum_i \log \mu_i$. 
We exclude ratios with $\mu_i \le 1$ (tied or zero distances, as with duplicate rows). However,
unlike in \citet{facco2017twonn}, we do not discard the largest ratios, which lower $\hat d$, 
so our convention reports the smaller inflation.

Additionally, following the protocol of~\citet{kataiwa2025tokenid}, we report the Levina--Bickel estimate with $k = 5$~\citep{levina2004mle},
the per-token value $m_i = (k-1) / \sum_{j<k} \log(r_{k,i} / r_{j,i})$
aggregated by its harmonic mean. We refer to it as MLE-5.

\textbf{Why a hub inflates them.}
Ratios near $\mu = 1$ push $\hat d$ up. Let $H$ be a hub, a set of points of norm at most 
$\sigma$, inside a cloud of points of norm near $\rho \gg \sigma$ with directions spread over $D$ dimensions.
 By concentration of measure two points of the cloud lie about $\rho\sqrt{2}$ apart, while a point of $H$ 
 lies within about $\rho$ of every point of the cloud. The nearest points of any $x$ 
 in the cloud are therefore in $H$, at nearly the same distance, so $r_2/r_1$ at $x$ is close to $1$.


A few hub points are enough to inflate the estimate, since each is
the nearest pair of thousands of others. 
Worse, the effect is strongest
where the estimate rests on few neighbors: exactly the case of the ID estimators. 

\begin{figure}[t]
\centering
\includegraphics[width=0.8\linewidth]{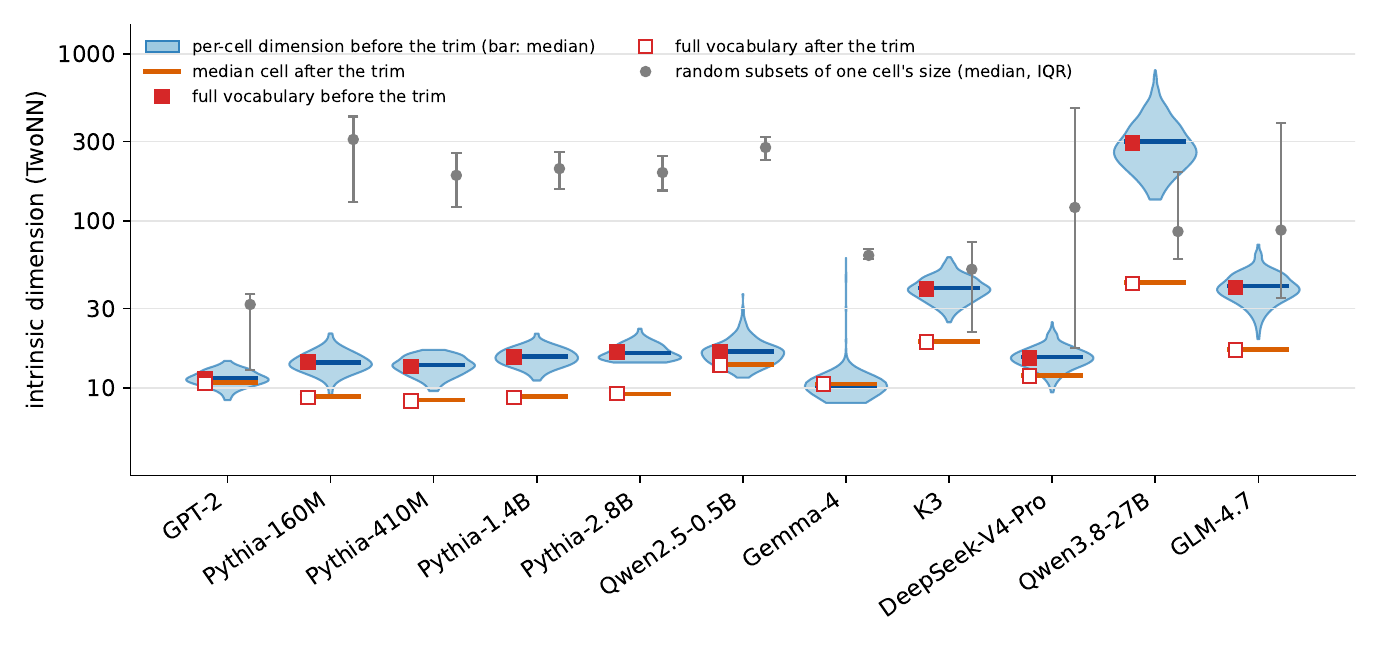}
\caption{Per-cell dimension before the trim (violins, medians as blue
bars), the median cell after the trim (orange bars), full-vocabulary TwoNN
before (filled squares) and after the trim (open squares), and TwoNN on
random subsets of one cell's size (grey, median and IQR).}
\label{fig:mixture}
\end{figure}

\textbf{Measuring the dimension of a table.} The full-vocabulary reading
takes the exact nearest neighbors of every row, giving TwoNN and MLE-5. 
In addition to the full-vocabulary reading, we report a per-region reading that splits the table into cells and reads each cell separately.
We splits the table with a random-projection
tree~\citep{dasgupta2008rptree} into cells of at most $800$ rows and
estimates the dimension per cell. We use cells rather than single tokens
because a single token gives one ratio, and one tie or one pair of
near-equal neighbors is enough to send that ratio anywhere. A cell is made of
hundreds of ratios, so no single token can drastically move its estimate. 
We tested this method on synthetic mixtures and it reads properly
 a planted dimension up to $d \approx 15$ and increasingly low above
it (Appendix~\ref{app:figs}).

\textbf{Which rows the trim removes.} Tokens are ranked by the norm of
their embedding row, and the trim removes the top decile (or the top $X\%$ for $X \in \{1, 5, 20\}$ in the sweep). 
Then, we re-runs the ID estimators with the removed rows no longer available as
neighbors. 

\section{Full-vocabulary and per-region readings}
\label{sec:field}
Table~\ref{tab:census} gives the readings before removing rows for thirteen models, 
including two larger Pythia tables added to reproduce
\citet{kataiwa2025tokenid}.

\begin{table}[t]
\centering
\caption{Thirteen tables before any trim: TwoNN, MLE-5, and the middle
half of the random-projection cells.}
\label{tab:census}
\scriptsize
\setlength{\tabcolsep}{3pt}
\begin{tabular}{lrrccc}
\toprule
Model & Params & $D$ & TwoNN & MLE-5 & Cells (IQR) \\
\midrule
GPT-2           & $0.12$B & $768$ & $11$ & $15$ & $11$--$12$ \\
Pythia-160M     & $0.16$B & $768$ & $14$ & $27$ & $13$--$16$ \\
Pythia-410M     & $0.41$B & $1024$ & $14$ & $25$ & $12$--$15$ \\
Pythia-1.4B     & $1.4$B & $2048$ & $15$ & $32$ & $14$--$17$ \\
Pythia-2.8B     & $2.8$B & $2560$ & $16$ & $34$ & $15$--$18$ \\
Pythia-6.9B     & $6.9$B & $4096$ & $30$ & $78$ & $28$--$33$ \\
Pythia-12B      & $12$B & $5120$ & $46$ & $122$ & $41$--$50$ \\
Qwen2.5-0.5B    & $0.49$B & $896$ & $16$ & $20$ & $15$--$18$ \\
Gemma-4         & $33$B & $5376$ & $10$ & $13$ & $9.8$--$11$ \\
K3              & $2.8$T & $7168$ & $39$ & $63$ & $36$--$45$ \\
DeepSeek-V4-Pro & $1.6$T & $7168$ & $15$ & $25$ & $14$--$17$ \\
Qwen3.8-27B     & $28$B & $5120$ & $294$ & $754$ & $241$--$388$ \\
GLM-4.7         & $358$B & $5120$ & $40$ & $71$ & $36$--$47$ \\
\bottomrule
\end{tabular}
\end{table}

On ten of the eleven the full-vocabulary estimated ID is small: $10$ to $40$
for TwoNN and $13$ to $71$ for MLE-5. Qwen3.8-27B is an exception, at
$294$ and $754$, that we explain by the fact that it has the deepest 
short-norm tail of all models. As we will see in Section~\ref{sec:geography}, 
the trim brings it to $42$ and $74$.
For Pythia family, MLE-5 climbs with the parameter count: $27$ to $122$, from 160M to
12B, reproducing the values of~\citet{kataiwa2025tokenid} on all six models. Similarly, 
TwoNN climbs from $14$ to $46$ with the parameter counts.

Figure~\ref{fig:mixture} shows for each model, the per-cell readings
before any trim (violins), the full-vocabulary reading (squares), and
TwoNN on random subsets of one cell's size (grey dots). On the inflated
tables the whole violin sits high: the
inflation is everywhere in the table, not in one region that could be
isolated. After the trim, the cells's ID estimation land on the trimmed
full-table value for every model, to within about $10\%$ (open squares
and orange bars). In other words, once the hub is gone, the table is 
uniform: every region reads the same dimension.



\section{A hub of short rows accounts for the inflation}
\label{sec:geography}


We removed the shortest-norm decile of each table and re-measured every
reading. 
We added three controls that each remove $10\%$: a random
$10\%$ (median of three seeds), the longest-norm decile, and the
farthest-from-centroid decile. We report in Table~\ref{tab:dust} the results with the median of the three seeds for the random control (every control arm in Table~\ref{tab:grid}). In addition, we report the purity of the
removed decile, i.e., the fraction of a removed token's ten nearest neighbors that
are also removed. We call a \emph{collapse} a trim that removes at least
$40\%$ of MLE-5.

\begin{table}[t]
\centering
\caption{The shortest-norm trim ($0\!\to\!10\%$) against a random $10\%$,
the untrimmed normalized reading, and the purity of the removed decile.}
\label{tab:dust}
\scriptsize
\setlength{\tabcolsep}{2.5pt}
\begin{tabular}{lccccc}
\toprule
Model & TwoNN $0\!\to\!10\%$ & MLE-5 $0\!\to\!10\%$ & Normalized & Random $0\!\to\!10\%$ & Purity \\
\midrule
GPT-2           & $11\!\to\!11$ & $15\!\to\!13$ & $12$ & $11\!\to\!12$ & $0.80$ \\
Pythia-160M     & $14\!\to\!8.8$ & $27\!\to\!10$ & $11$ & $14\!\to\!15$ & $0.94$ \\
Pythia-410M     & $14\!\to\!8.4$ & $25\!\to\!9.5$ & $9.8$ & $14\!\to\!14$ & $0.97$ \\
Pythia-1.4B     & $15\!\to\!8.8$ & $32\!\to\!9.9$ & $10$ & $15\!\to\!16$ & $0.98$ \\
Pythia-2.8B     & $16\!\to\!9.3$ & $34\!\to\!10$ & $10$ & $16\!\to\!17$ & $0.98$ \\
Qwen2.5-0.5B    & $16\!\to\!14$ & $20\!\to\!15$ & $15$ & $16\!\to\!17$ & $0.90$ \\
Gemma-4         & $10\!\to\!11$ & $13\!\to\!13$ & $13$ & $10\!\to\!11$ & $0.28$ \\
K3              & $39\!\to\!19$ & $63\!\to\!20$ & $17$ & $39\!\to\!41$ & $0.99$ \\
DeepSeek-V4-Pro & $15\!\to\!12$ & $25\!\to\!14$ & $13$ & $15\!\to\!16$ & $0.99$ \\
Qwen3.8-27B     & $294\!\to\!42$ & $754\!\to\!74$ & $34$ & $294\!\to\!320$ & $1.00$ \\
GLM-4.7         & $40\!\to\!17$ & $71\!\to\!18$ & $90$ & $40\!\to\!42$ & $1.00$ \\
\bottomrule
\end{tabular}
\end{table}

\textbf{Eight collapses.} On eight of the eleven tables the trim
removes $41$--$90\%$ of the MLE-5 reading and $22$--$86\%$
of the TwoNN reading, while the controls move the readings the other way:
removing a random $10\%$ changes MLE-5 by $+6$ to $+10\%$ and the
longest-norm decile by $-4$ to $+39\%$ (Table~\ref{tab:grid}).
By removing any 10\% the table is sparser, and as one would expect, the 
estimate drifts up. The opposite happens for the trim of the shortest decile: readings collapse, precisely because it is the hub that inflates them in the first place. The three tables that move less, and the most inflated one, are examined in Appendix~\ref{app:cases}. 

We reproduce exactly the published values of~\citet{kataiwa2025tokenid} for Pythia, including the rise with the parameter count. 
However, after the trim, the readings are flat with the parameter count. The rise was a statistical artifact. Our interpretation is that before the trim, 
the hub inflates the readings due to the geometric effect described in Section~\ref{sec:instruments},
  but once the hub is gone, a token's neighbors are other trained tokens, and their spacing reflects 
 how training arranged the vocabulary: a characteristic of the model's architecture and the language itself, not the training.
 The trained rows occupy a low-dimensional set, and that set is embedded in a wider space at 12B than at 160M without becoming wider itself. 
 The extra width is unused, which is  the {\it redundancy} mentioned in~\citet{kataiwa2025tokenid}: the redundancy grows with the parameter count, the dimension does not.

\textbf{The effect is a switch.} A few hundred rows are enough to fully inflate the estimate. 
Removing only the shortest $1\%$ of a table already collapses the reading as shown in Table~\ref{tab:sweep}).
Furthermore, returning a random quarter of the removed decile restores the untrimmed value in full (Pythia-160M $10 \to 26$,
GLM-4.7 $18 \to 78$, with three subsets giving the same results within $0.5$ (Table~\ref{tab:reinsert}, Figure~\ref{fig:sandbox} in Appendix~\ref{app:figs}). Returning the
padding rows alone does the same (Table~\ref{tab:padding}), indicating an artifact of the tokenizer itself.

\begin{figure}[t]
\centering
\includegraphics[width=0.5\linewidth]{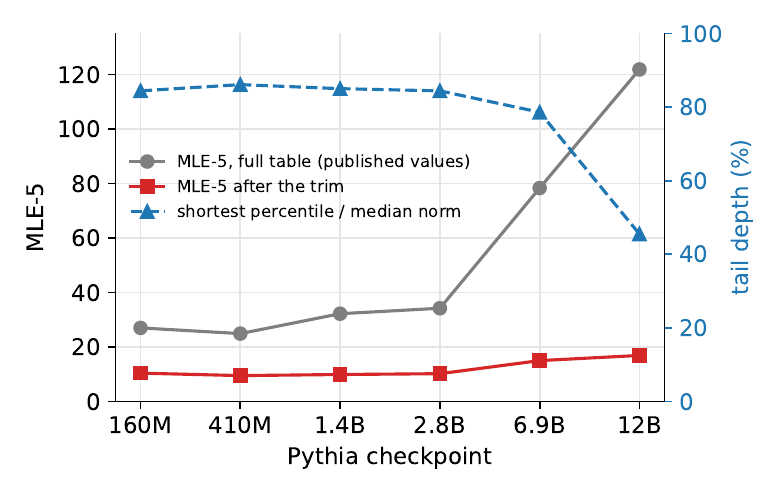}
\caption{Pythia across parameter counts: MLE-5 before and after the trim,
and the depth of the norm tail.}
\label{fig:ladder}
\end{figure}

\textbf{The norm tail predicts the inflation.} 
We define the norm tail as the 1st percentile of the norms divided by the median norm.
The tail deepens at 6.9B and 12B, exactly where the published values from~\citet{kataiwa2025tokenid} jump (Figure~\ref{fig:ladder}, Table~\ref{tab:protocol}). 
The norms predict the inflation: the further a checkpoint's shortest rows sit below its median norm, 
the higher a ID estimator will read. Gemma-4 has no short tail and the ID estimation is not inflated, Qwen3.8-27B has the deepest and reads highest.
This can possibly be used as a diagnostic during training: a deep tail flags the tokens that never received gradients, and it warns that 
any neighbor-based reading of that table must be trimmed or normalized before it can be compared to anything. We let this for future work.

\textbf{Equalizing the norms removes the hub.} Because the distortion comes
from short rows near the origin, we scaled every row to unit length instead
of deleting them. In other words, we replaced the Euclidean distance with the cosine distance.
Not only it is standard in the representation-learning literature precisely because of the concentration of measure in high dimensions,
but it is also the simplest of the hubness reductions surveyed by \citet{feldbauer2019hubness}. As a result, Pythia tables read between $10$, and $17$
for MLE-5 for all parameter counts, when the raw tables read $27$ to $122$
(Table~\ref{tab:dust}, Table~\ref{tab:cosine}). The dimension estimations are not inflated anymore despite the fact we did not remove any rows.

The normalization works as well on large models: normalized, K3 reads $17$ ($20$
trimmed), DeepSeek-V4-Pro $13$ ($14$ trimmed), and Qwen3.8-27B $34$ ($74$
trimmed). On Qwen3.8-27B no trim reaches the normalized reading (between $754$
and $62$ at $0$ to $20\%$ against $34$): on the table with
the deepest tail, normalization removes more than deletion does, and why
is an open question. GLM-4.7 is the one exception but the explanation is simple: the model 
has $485$ near-zero rows and they all share one
direction, so on the unit sphere they become one point that supplies all
five neighbors of the tokens near it, which mechanically lowers the ID estimate.


\section{The per-token tail and the under-trained tokens are the hub}
\label{sec:defense}

\begin{figure}[t]
\centering
\includegraphics[width=0.95\linewidth]{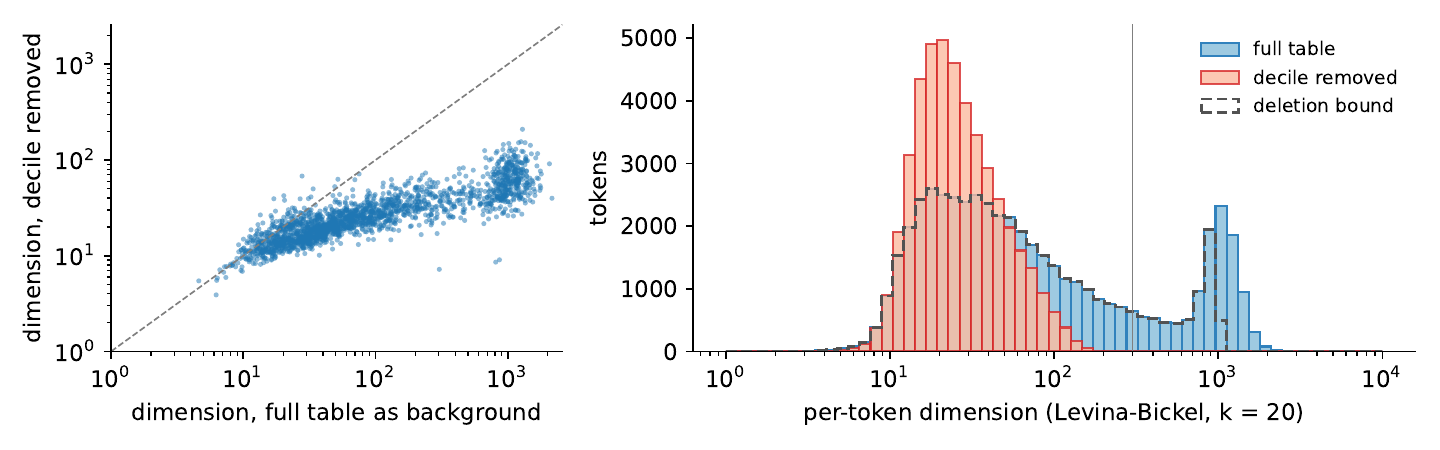}
\caption{Per-token dimension on GPT-2 before and after the trim. Left:
each of $2{,}000$ kept tokens against itself, full table as background
($x$) and shortest decile removed ($y$). Right: the distribution of all
tokens (log axis). The dashed outline is the deletion bound, the lowest
the distribution could sit if the trim had only deleted the highest
readers.}
\label{fig:pertoken}
\end{figure}

 \textbf{The per-token tail is the hub.} \citet{robinson2025manifold} measure a dimension 
 at every token of GPT-2, find a heavy tail (median $389$), and read it as a violation of the 
 manifold hypothesis. We first reproduce a tail using MLE-20 per-token: the median 
 is $52$, and $22\%$ of tokens read above $300$. Removing the shortest-norm decile ends it: the median
  falls to $24$ and no token stays above $300$ (Figure~\ref{fig:pertoken}, right). 
  The comparison is not circular, because the trim sees only each row's norm. 
  The removed rows themselves have a low ID (median $14$), so the hub does not carry the tail, it induces 
  it in the others, and the tokens it inflates most are the longest rows ($54\%$ of the longest
   quarter's ID is above $300$, against $4\%$ of the shortest), as the geometry predicts. 
   The tail does not disappear because high dimensional tokens were deleted. It is the mechanism 
   seen one token at a time: $93.5\%$ of the tokens the trim never touches have then a lower ID estimate once 
   the hub is gone, and the higher a token read before, the further its ID falls (Spearman $0.88$, Figure~\ref{fig:pertoken}, left). 
   Normalizing the rows instead of removing any gives the same per-token median ($24$).

On top of our approach, we then ran their released test on $4{,}000$ tokens before and after the trim. 
At its small-radius window the median is $507$\footnote{Our $507$ against their $389$ is the sample and the window: 
tokens drawn from every row read $490$, and we used their code's default window rather than the paper's.} before the trim and $41$ after, and the share of 
tokens above $300$ falls from $82\%$ to none: their heavy tail is the hub. At a large-radius window 
the reading barely moves ($46$ to $57$), because far neighbors were never hub rows. Note that their formal rejections of 
the manifold hypothesis do not move either: $3$ tokens out of$4{,}000$ are marked as singularities before, $2$ after. So at small radius their instrument 
measures the hub, at large radius it measures the table, and the tail disappears with the hub while the few 
genuinely rejected tokens remain.

\textbf{An external detector.} Magikarp~\citep{land2024magikarp} ranks tokens 
by a per-architecture indicator and verifies its candidates by prompting the model. 
On GPT-2 its indicator is the cosine distance to the mean row, not the norm,
 so we can fairly compare if the under-trained tokens that they detect are the hub.
The lower deciles of their ranking share only $6\%$  with our decile.
However, we checked the $97$ tokens that they verified by prompting, and $50$ of them fall in our decile.

 On Pythia-6.9B, its indicator is also the embedding norm itself, therefore,
 so agreement there proves nothing, but all $36$ verified tokens fall in our decile.
 So the verified under-trained tokens seem to concentrate in the hub.

\textbf{Are they under-trained?} The hub is the population the
detectors select, and its extreme members were never trained: the padding
rows, GLM-4.7's $128$ zero rows, and the verified tokens above. For the
decile as a whole the evidence is weaker. Against Pythia's initial checkpoint, the removed rows of Pythia-160M and 410M have moved
as far as every other row (cosine $0.02$ to where they started), so they
were updated during training, and what distinguishes them is that they
ended short. On Pythia at least, beyond the extreme members, the hub means less trained
rather than untrained. Conversely, short rows are notrare rows in general: norm and
corpus count correlate positively on seven tables and negatively on GPT-2
and DeepSeek-V4-Pro ($-0.46$, $-0.29$), where the shortest rows are
frequent tokens (Table~\ref{tab:protocol}).

\section{Discussion and conclusion}
\label{sec:discussion}
A token-embedding table holds a hub of short rows near its origin, and
in a high-dimensional table that hub is the nearest pair of most trained
tokens at nearly equal distances, so nearest-neighbor estimators read a
dimension the table does not have. The inflation is an artifact of
Euclidean neighbor-ratio estimators on a table whose rows differ in
length, not a property of the tables, and reading the table under cosine
distance removes it everywhere. 
What remains is close to one number: the regions of a table read within about $10\%$ of
each other, and the full-vocabulary estimate is
their harmonic mean.

We reproduced the values of \citet{kataiwa2025tokenid}, but the growth with the parameter 
count did not survive the hub removal: it tracks the depth of the norm tail, and after the 
trim the six models read $9.5$ to $17$. For \citet{robinson2025manifold}, the heavy tail
 comes from the hub as well: after the trim, the median falls from $507$ to $41$.
  Their test still rejects the manifold hypothesis on $3$ tokens before the trim and $2$ after, 
  and we do not explain these tokens.

In practice, we recommend to report the estimate before and after removing 
the shortest decile, or on normalized rows: if the two disagree, the embedding has a hub. 
Checking the norm histogram is a cheap way to tell in advance if the raw estimate can be trusted.

Among the remaining open questions, we do not know why a table grows a tail. The three models with tied embeddings are 
the three without a tail, so our hypothesis
 is that tied weights update every row at every step, while an untied row is only updated when its token occurs. 
 Also, we could not prove that the hub is under-trained: its extreme members were never trained, but on Pythia the 
 removed rows moved as much as the others during training. 

Future work will focus on the unembedding matrix, which we left aside; on following hub tokens through the residual stream to further characterize them, and on whether tied and untied models differ in ID estimation beyond the depth of their tails.

\bibliography{references}

@article{facco2017twonn,
  title={Estimating the intrinsic dimension of datasets by a minimal neighborhood information},
  author={Facco, Elena and d'Errico, Maria and Rodriguez, Alex and Laio, Alessandro},
  journal={Scientific Reports}, volume={7}, pages={12140}, year={2017}
}

@inproceedings{levina2004mle,
  title={Maximum Likelihood Estimation of Intrinsic Dimension},
  author={Levina, Elizaveta and Bickel, Peter J.},
  booktitle={Advances in Neural Information Processing Systems (NeurIPS)}, year={2004}
}

@inproceedings{ansuini2019intrinsic,
  title={Intrinsic dimension of data representations in deep neural networks},
  author={Ansuini, Alessio and Laio, Alessandro and Macke, Jakob H. and Zoccolan, Davide},
  booktitle={Advances in Neural Information Processing Systems (NeurIPS)}, year={2019}
}

@inproceedings{valeriani2023geometry,
  title={The geometry of hidden representations of large transformer models},
  author={Valeriani, Lucrezia and Doimo, Diego and Cuturello, Francesca and Laio, Alessandro and Ansuini, Alessio and Cazzaniga, Alberto},
  booktitle={Advances in Neural Information Processing Systems (NeurIPS)}, year={2023}
}

@inproceedings{houle2017lid,
  title={Local Intrinsic Dimensionality {I}: An Extreme-Value-Theoretic Foundation for Similarity Applications},
  author={Houle, Michael E.},
  booktitle={International Conference on Similarity Search and Applications (SISAP)}, year={2017}
}

@inproceedings{amsaleg2015lid,
  title={Estimating Local Intrinsic Dimensionality},
  author={Amsaleg, Laurent and Chelly, Oussama and Furon, Teddy and Girard, St{\'e}phane and Houle, Michael E. and Kawarabayashi, Ken-ichi and Nett, Michael},
  booktitle={ACM SIGKDD International Conference on Knowledge Discovery and Data Mining (KDD)}, year={2015}
}

@article{allegra2020hidalgo,
  title={Data segmentation based on the local intrinsic dimension},
  author={Allegra, Michele and Facco, Elena and Denti, Francesco and Laio, Alessandro and Mira, Antonietta},
  journal={Scientific Reports}, volume={10}, pages={16449}, year={2020}
}

@inproceedings{robinson2025manifold,
  title={Token Embeddings Violate the Manifold Hypothesis},
  author={Robinson, Michael and Dey, Sourya and Chiang, Tony},
  booktitle={Advances in Neural Information Processing Systems (NeurIPS)}, year={2025},
  note={arXiv:2504.01002}
}

@article{kataiwa2025tokenid,
  title={Measuring Intrinsic Dimension of Token Embeddings},
  author={Kataiwa, Takuya and Hakaze, Cho and Ohki, Tetsushi},
  journal={arXiv preprint arXiv:2503.02142}, year={2025}
}

@inproceedings{ethayarajh2019contextual,
  title={How Contextual are Contextualized Word Representations? Comparing the Geometry of BERT, ELMo, and GPT-2 Embeddings},
  author={Ethayarajh, Kawin},
  booktitle={Proceedings of the 2019 Conference on Empirical Methods in Natural Language Processing (EMNLP-IJCNLP)},
  year={2019}
}

@inproceedings{timkey2021bark,
  title={All Bark and No Bite: Rogue Dimensions in Transformer Language Models Obscure Representational Quality},
  author={Timkey, William and van Schijndel, Marten},
  booktitle={Proceedings of the 2021 Conference on Empirical Methods in Natural Language Processing (EMNLP)},
  year={2021}
}

@inproceedings{land2024magikarp,
  title={Fishing for Magikarp: Automatically Detecting Under-trained Tokens in Large Language Models},
  author={Land, Sander and Bartolo, Max},
  booktitle={Proceedings of the 2024 Conference on Empirical Methods in Natural Language Processing (EMNLP)},
  pages={11631--11646}, year={2024}
}

@inproceedings{dasgupta2008rptree,
  title={Random projection trees and low dimensional manifolds},
  author={Dasgupta, Sanjoy and Freund, Yoav},
  booktitle={Proceedings of the 40th Annual ACM Symposium on Theory of Computing (STOC)},
  pages={537--546}, year={2008}
}

@inproceedings{gao2019degeneration,
  title={Representation Degeneration Problem in Training Natural Language Generation Models},
  author={Gao, Jun and He, Di and Tan, Xu and Qin, Tao and Wang, Liwei and Liu, Tie-Yan},
  booktitle={International Conference on Learning Representations (ICLR)}, year={2019},
  note={arXiv:1907.12009}
}

@inproceedings{mu2018allbutthetop,
  title={All-but-the-Top: Simple and Effective Postprocessing for Word Representations},
  author={Mu, Jiaqi and Viswanath, Pramod},
  booktitle={International Conference on Learning Representations (ICLR)}, year={2018},
  note={arXiv:1702.01417}
}

@inproceedings{rudman2022isoscore,
  title={IsoScore: Measuring the Uniformity of Embedding Space Utilization},
  author={Rudman, William and Gillman, Nate and Rayne, Taylor and Eickhoff, Carsten},
  booktitle={Findings of the Association for Computational Linguistics: ACL 2022},
  pages={3325--3339}, year={2022}
}

@misc{rumbelow2023magikarp,
  title={{SolidGoldMagikarp} (plus, prompt generation)},
  author={Rumbelow, Jessica and Watkins, Matthew},
  howpublished={LessWrong / AI Alignment Forum, technical report}, year={2023}
}

@article{denti2022gride,
  title={The generalized ratios intrinsic dimension estimator},
  author={Denti, Francesco and Doimo, Diego and Laio, Alessandro and Mira, Antonietta},
  journal={Scientific Reports}, volume={12}, pages={20005}, year={2022}
}

@misc{mackay2005correction,
  title={Comments on `Maximum Likelihood Estimation of Intrinsic Dimension' by {E. Levina} and {P. Bickel}},
  author={MacKay, David J. C. and Ghahramani, Zoubin},
  howpublished={Unpublished note, \url{http://www.inference.org.uk/mackay/dimension/}}, year={2005}
}

@inproceedings{aghajanyan2021intrinsic,
  title={Intrinsic Dimensionality Explains the Effectiveness of Language Model Fine-Tuning},
  author={Aghajanyan, Armen and Gupta, Sonal and Zettlemoyer, Luke},
  booktitle={Proceedings of the 59th Annual Meeting of the Association for Computational Linguistics (ACL)},
  year={2021}, note={arXiv:2012.13255}
}

@inproceedings{razzhigaev2024shape,
  title={The Shape of Learning: Anisotropy and Intrinsic Dimensions in Transformer-Based Models},
  author={Razzhigaev, Anton and Mikhalchuk, Matvey and Goncharova, Elizaveta and Oseledets, Ivan and Dimitrov, Denis and Kuznetsov, Andrey},
  booktitle={Findings of the Association for Computational Linguistics: EACL 2024},
  year={2024}, note={arXiv:2311.05928}
}

@inproceedings{wu2026unembedding,
  title={Your UnEmbedding Matrix is Secretly a Feature Lens for Text Embeddings},
  author={Wu, Songhao and Chen, Zhongxin and Liu, Yuxuan and Cui, Heng and Li, Cong and Yan, Rui},
  booktitle={Proceedings of the 32nd ACM SIGKDD Conference on Knowledge Discovery and Data Mining (KDD), V.2},
  year={2026},
  doi={10.1145/3770855.3818204},
  note={arXiv:2606.07502}
}

@article{radovanovic2010hubs,
  title   = {Hubs in Space: Popular Nearest Neighbors in High-Dimensional Data},
  author  = {Radovanovi{\'c}, Milo{\v{s}} and Nanopoulos, Alexandros and Ivanovi{\'c}, Mirjana},
  journal = {Journal of Machine Learning Research},
  volume  = {11},
  pages   = {2487--2531},
  year    = {2010}
}

@article{schnitzer2012mutual,
  title   = {Local and Global Scaling Reduce Hubs in Space},
  author  = {Schnitzer, Dominik and Flexer, Arthur and Schedl, Markus and Widmer, Gerhard},
  journal = {Journal of Machine Learning Research},
  volume  = {13},
  pages   = {2871--2902},
  year    = {2012}
}

@article{feldbauer2019hubness,
  title   = {A Comprehensive Empirical Comparison of Hubness Reduction in High-Dimensional Spaces},
  author  = {Feldbauer, Roman and Flexer, Arthur},
  journal = {Knowledge and Information Systems},
  volume  = {59},
  pages   = {137--166},
  year    = {2019}
}

@inproceedings{dinu2015improving,
  title     = {Improving Zero-Shot Learning by Mitigating the Hubness Problem},
  author    = {Dinu, Georgiana and Lazaridou, Angeliki and Baroni, Marco},
  booktitle = {International Conference on Learning Representations, Workshop Track},
  year      = {2015},
  note      = {arXiv:1412.6568}
}

\appendix
\section{Complete trim results}
\label{app:grid}
Table~\ref{tab:grid} reports every arm of the trim design for all eleven
tables, on the full vocabulary. For each of the two readings, TwoNN and
the Levina--Bickel estimate with $k = 5$ neighbors aggregated by harmonic
mean (MLE-5), the first column repeats the trim of
Table~\ref{tab:dust} and the next three give the reading after removing a
random $10\%$ (median of three seeds), the longest-norm decile, and the
farthest-from-centroid decile. In every trimmed arm the removed rows are
excluded as neighbors and the estimate is taken over the remaining rows.

Table~\ref{tab:padding} tests the padding rows alone and the shortest decile
with the padding rows left in, and Table~\ref{tab:cosine} repeats the
trim on row-normalized embeddings.
Table~\ref{tab:sweep} gives both readings after removing the shortest $1$,
$5$, $10$, and $20\%$, Table~\ref{tab:reinsert} the re-insertion sweep on
the seven collapsing tables, returning a random quarter, half, and three
quarters of the removed decile, and Table~\ref{tab:curve} the TwoNN
reading on random subsamples of four tables, the sampling-density effect
of Section~\ref{sec:field}.

\begin{table}[h]
\centering
\caption{Every trim arm on the full vocabulary, for both readings. Random
is the median of three seeds. Long removes the longest-norm decile. The
farthest-from-centroid decile reads within one unit of Long on every
table and is omitted.}
\label{tab:grid}
\footnotesize
\setlength{\tabcolsep}{3pt}
\begin{tabular}{lccc ccc}
\toprule
& \multicolumn{3}{c}{TwoNN} & \multicolumn{3}{c}{MLE-5} \\
\cmidrule(lr){2-4} \cmidrule(lr){5-7}
Model & Trim $0\!\to\!10\%$ & Random & Long & Trim $0\!\to\!10\%$ & Random & Long \\
\midrule
GPT-2           & $11\!\to\!11$ & $12$ & $11$ & $15\!\to\!13$ & $15$ & $14$ \\
Pythia-160M     & $14\!\to\!8.8$ & $15$ & $16$ & $27\!\to\!10$ & $29$ & $32$ \\
Pythia-410M     & $14\!\to\!8.4$ & $14$ & $15$ & $25\!\to\!9.5$ & $27$ & $28$ \\
Pythia-1.4B     & $15\!\to\!8.8$ & $16$ & $17$ & $32\!\to\!9.9$ & $35$ & $37$ \\
Pythia-2.8B     & $16\!\to\!9.3$ & $17$ & $18$ & $34\!\to\!10$ & $37$ & $39$ \\
Qwen2.5-0.5B    & $16\!\to\!14$ & $17$ & $16$ & $20\!\to\!15$ & $20$ & $19$ \\
Gemma-4         & $10\!\to\!11$ & $11$ & $10$ & $13\!\to\!13$ & $13$ & $13$ \\
K3              & $39\!\to\!19$ & $41$ & $45$ & $63\!\to\!20$ & $67$ & $88$ \\
DeepSeek-V4-Pro & $15\!\to\!12$ & $16$ & $15$ & $25\!\to\!14$ & $27$ & $24$ \\
Qwen3.8-27B     & $294\!\to\!42$ & $320$ & $466$ & $754\!\to\!74$ & $804$ & $1004$ \\
GLM-4.7         & $40\!\to\!17$ & $42$ & $47$ & $71\!\to\!18$ & $77$ & $88$ \\
\bottomrule
\end{tabular}
\end{table}

\begin{table}[h]
\centering
\caption{Trim sweep on the full vocabulary: MLE-5 and, in parentheses,
TwoNN after removing the shortest $X\%$.}
\label{tab:sweep}
\footnotesize
\begin{tabular}{lccccc}
\toprule
Model & $0\%$ & $1\%$ & $5\%$ & $10\%$ & $20\%$ \\
\midrule
GPT-2           & $15$ ($11$) & $15$ ($11$) & $12$ ($11$) & $13$ ($11$) & $14$ ($11$) \\
Pythia-160M     & $27$ ($14$) & $11$ ($9.1$) & $11$ ($9.0$) & $10$ ($8.8$) & $10$ ($8.6$) \\
Pythia-410M     & $25$ ($14$) & $9.9$ ($8.6$) & $9.6$ ($8.5$) & $9.5$ ($8.4$) & $9.3$ ($8.3$) \\
Pythia-1.4B     & $32$ ($15$) & $10$ ($9.0$) & $10$ ($8.9$) & $9.9$ ($8.8$) & $9.7$ ($8.7$) \\
Pythia-2.8B     & $34$ ($16$) & $11$ ($9.5$) & $10$ ($9.3$) & $10$ ($9.3$) & $10$ ($9.2$) \\
Qwen2.5-0.5B    & $20$ ($16$) & $20$ ($16$) & $16$ ($14$) & $15$ ($14$) & $15$ ($14$) \\
Gemma-4         & $13$ ($10$) & $12$ ($10$) & $12$ ($10$) & $13$ ($11$) & $13$ ($11$) \\
K3              & $63$ ($39$) & $28$ ($23$) & $22$ ($20$) & $20$ ($19$) & $19$ ($18$) \\
DeepSeek-V4-Pro & $25$ ($15$) & $16$ ($13$) & $15$ ($12$) & $14$ ($12$) & $15$ ($12$) \\
Qwen3.8-27B     & $754$ ($294$) & $223$ ($123$) & $90$ ($50$) & $74$ ($42$) & $62$ ($37$) \\
GLM-4.7         & $71$ ($40$) & $31$ ($24$) & $20$ ($18$) & $18$ ($17$) & $17$ ($16$) \\
\bottomrule
\end{tabular}
\end{table}

\begin{table}[h]
\centering
\caption{The re-insertion sweep on the full vocabulary: MLE-5 and, in
parentheses, TwoNN after returning a random fraction of the removed decile
to the trimmed table.}
\label{tab:reinsert}
\footnotesize
\begin{tabular}{lccccc}
\toprule
Model & $0\%$ & $25\%$ & $50\%$ & $75\%$ & $100\%$ \\
\midrule
Pythia-160M     & $10$ ($8.8$) & $26$ ($14$) & $27$ ($14$) & $27$ ($14$) & $27$ ($14$) \\
Pythia-410M     & $9.5$ ($8.4$) & $24$ ($13$) & $24$ ($13$) & $25$ ($13$) & $25$ ($14$) \\
Pythia-1.4B     & $9.9$ ($8.8$) & $31$ ($15$) & $32$ ($15$) & $32$ ($15$) & $32$ ($15$) \\
Pythia-2.8B     & $10$ ($9.3$) & $33$ ($16$) & $34$ ($16$) & $34$ ($16$) & $34$ ($16$) \\
K3              & $20$ ($19$) & $64$ ($39$) & $64$ ($39$) & $63$ ($39$) & $63$ ($39$) \\
DeepSeek-V4-Pro & $14$ ($12$) & $31$ ($17$) & $28$ ($16$) & $26$ ($16$) & $25$ ($15$) \\
GLM-4.7         & $18$ ($17$) & $78$ ($40$) & $75$ ($40$) & $72$ ($40$) & $71$ ($40$) \\
\bottomrule
\end{tabular}
\end{table}

\begin{table}[h]
\centering
\caption{TwoNN on random subsamples of the untrimmed table, median over
five independent draws with the range in brackets, and on the full
vocabulary.}
\label{tab:curve}
\footnotesize
\begin{tabular}{lccccc}
\toprule
Model & $1$k & $4$k & $16$k & $64$k & Full \\
\midrule
Pythia-160M     & $161$ [$69$--$210$] & $78$ [$69$--$94$] & $28$ [$26$--$28$] & -- & $14$ \\
K3              & $47$ [$20$--$134$] & $172$ [$72$--$186$] & $127$ [$123$--$135$] & $58$ [$57$--$58$] & $39$ \\
DeepSeek-V4-Pro & $334$ [$48$--$410$] & $148$ [$133$--$163$] & $49$ [$48$--$52$] & $20$ [$20$--$20$] & $15$ \\
GLM-4.7         & $339$ [$16$--$4{,}470$] & $393$ [$310$--$428$] & $161$ [$152$--$174$] & $60$ [$58$--$61$] & $40$ \\
\bottomrule
\end{tabular}
\end{table}

\begin{table}[h]
\centering
\caption{Partition robustness on all eleven tables, exact neighbors: the
untrimmed median cell over five random-projection seeds (cap $800$, median
and range), the trimmed median cell (seed $0$), and the number of cells.}
\label{tab:robust}
\footnotesize
\begin{tabular}{lccc}
\toprule
Model & Seeds $0$--$4$ & Trimmed & Cells \\
\midrule
GPT-2           & $11$ [$11$--$12$] & $11$ & $64$ \\
Pythia-160M     & $14$ [$14$--$14$] & $8.9$ & $64$ \\
Pythia-410M     & $14$ [$13$--$14$] & $8.5$ & $64$ \\
Pythia-1.4B     & $15$ [$15$--$16$] & $8.9$ & $64$ \\
Pythia-2.8B     & $16$ [$16$--$17$] & $9.2$ & $64$ \\
Qwen2.5-0.5B    & $16$ [$16$--$17$] & $14$ & $256$ \\
Gemma-4         & $10$ [$10$--$10$] & $11$ & $512$ \\
K3              & $40$ [$40$--$40$] & $19$ & $256$ \\
DeepSeek-V4-Pro & $15$ [$15$--$15$] & $12$ & $256$ \\
Qwen3.8-27B     & $303$ [$300$--$312$] & $43$ & $512$ \\
GLM-4.7         & $41$ [$40$--$41$] & $17$ & $256$ \\
\bottomrule
\end{tabular}
\end{table}


\begin{table}[h]
\centering
\caption{The padding test, MLE-5 on the full vocabulary. Rows: the number
of rows beyond the tokenizer's vocabulary plus zero-norm rows, then
MLE-5 untrimmed, after removing only those rows, after removing the
shortest decile but leaving those rows in, and after the full trim.}
\label{tab:padding}
\footnotesize
\begin{tabular}{lccccc}
\toprule
Model & Rows & Untrimmed & Rows only & Trim, rows kept & Trim \\
\midrule
Pythia-160M     & $27$ & $27$ & $27$ & $26$ & $10$ \\
Pythia-410M     & $27$ & $25$ & $25$ & $24$ & $9.5$ \\
Pythia-1.4B     & $27$ & $32$ & $32$ & $30$ & $9.9$ \\
Pythia-2.8B     & $27$ & $34$ & $34$ & $32$ & $10$ \\
Pythia-6.9B     & $155$ & $78$ & $76$ & $77$ & $15$ \\
Pythia-12B      & $411$ & $122$ & $116$ & $124$ & $17$ \\
Qwen2.5-0.5B    & $271$ & $20$ & $20$ & $20$ & $15$ \\
Qwen3.8-27B     & $243$ & $754$ & $754$ & $475$ & $74$ \\
GLM-4.7         & $315$ & $71$ & $71$ & $84$ & $18$ \\
\bottomrule
\end{tabular}
\end{table}

\begin{table}[h]
\centering
\caption{The metric control: the trim ($0\!\to\!10\%$) on raw rows,
on row-normalized rows.}
\label{tab:cosine}
\footnotesize
\setlength{\tabcolsep}{3.5pt}
\begin{tabular}{lcc cc}
\toprule
& \multicolumn{2}{c}{MLE-5} & \multicolumn{2}{c}{TwoNN} \\
\cmidrule(lr){2-3} \cmidrule(lr){4-5}
Model & Raw & Normalized & Raw & Normalized \\
\midrule
GPT-2           & $15\!\to\!13$ & $12\!\to\!12$ & $11\!\to\!11$ & $10\!\to\!10$ \\
Pythia-160M     & $27\!\to\!10$ & $11\!\to\!10$ & $14\!\to\!8.8$ & $9.1\!\to\!8.8$ \\
Pythia-410M     & $25\!\to\!9.5$ & $9.8\!\to\!9.5$ & $14\!\to\!8.4$ & $8.7\!\to\!8.4$ \\
Pythia-1.4B     & $32\!\to\!9.9$ & $10\!\to\!9.9$ & $15\!\to\!8.8$ & $9.0\!\to\!8.8$ \\
Pythia-2.8B     & $34\!\to\!10$ & $10\!\to\!10$ & $16\!\to\!9.3$ & $9.4\!\to\!9.2$ \\
Pythia-6.9B     & $78\!\to\!15$ & $15\!\to\!15$ & $30\!\to\!12$ & $13\!\to\!12$ \\
Pythia-12B      & $122\!\to\!17$ & $17\!\to\!17$ & $46\!\to\!14$ & $14\!\to\!14$ \\
Qwen2.5-0.5B    & $20\!\to\!15$ & $15\!\to\!14$ & $16\!\to\!14$ & $14\!\to\!13$ \\
Gemma-4         & $13\!\to\!13$ & $13\!\to\!13$ & $10\!\to\!11$ & $10\!\to\!11$ \\
K3              & $63\!\to\!20$ & $17\!\to\!16$ & $39\!\to\!19$ & $17\!\to\!16$ \\
DeepSeek-V4-Pro & $25\!\to\!14$ & $13\!\to\!13$ & $15\!\to\!12$ & $11\!\to\!11$ \\
Qwen3.8-27B     & $754\!\to\!74$ & $34\!\to\!34$ & $294\!\to\!42$ & $25\!\to\!25$ \\
GLM-4.7         & $71\!\to\!18$ & $90\!\to\!17$ & $40\!\to\!17$ & $15\!\to\!16$ \\
\bottomrule
\end{tabular}
\end{table}

\section{Case studies: the negative controls and the extreme table}
\label{app:cases}
\textbf{The negative controls.} Gemma-4 is clean: trimming changes nothing
($12.6 \to 12.8$ for MLE-5), even though its $262$k vocabulary is the
largest in the survey, because its rows all have the same norm by construction. 
GPT-2 moves only mildly
($14.7 \to 12.8$), and its shortest rows are frequent tokens rather than
rare ones (Section~\ref{sec:defense}). If the procedure manufactured
collapses, these two would have collapsed too which is why we consider them as negative controls.

\textbf{Qwen2.5-0.5B collapses partway.} The trim takes the reading from $20$ to $15$ and normalization gives the same $15$, 
a quarter of the reading against half or more on the collapsing tables. 
Like the other tied tables it has no row below half the median norm. A shallow tail, a small inflation.

\textbf{Qwen3.8-27B.}  Before the trim this is the most inflated table in the survey ($294$ for TwoNN, $754$ for MLE-5), 
and it has the deepest short tail, with $3{,}924$ rows below half the median norm and $1{,}224$ below a quarter. 
The removed decile supplies the nearest neighbor of $99.7\%$ of the other rows. 
The trim removes $86$ and $90\%$ of the two readings, and the median cell falls from $300$ to $43$, 
so the regions collapse together with the pooled number. The sweep reads $754$, $223$, $90$, and $74$ at $0$, $1$, $5$, 
and $10\%$, so most of the hub sits in the shortest few percent, and the decile is the setting of our instrument rather 
than the size of the hub. The vocabulary has $248$k rows, includes multimodal tokens, and the proxy corpus sees a
 quarter of it, so the frequency numbers mean little on this table. The norms and the neighbors do not depend on the corpus.

\section{Reproducibility}
\label{app:repro}
Table~\ref{tab:protocol} lists the embedding
tying, stored precision, and the depth of the norm tail of every
checkpoint, and Table~\ref{tab:checkpoints} its row counts. 
All tables are the input embedding matrix
of the released checkpoint, analyzed in float32. Nearest neighbors are
exact, computed over the full vocabulary, and every trimmed arm removes
its rows both as points and as neighbors. Rows whose first neighbor
distance is zero are excluded from both readings (GLM-4.7's $128$
zero-norm rows and $127$ duplicate pairs, two duplicate pairs on
Qwen2.5-0.5B). All seeds are fixed and no component is trained. Token
frequencies are counted on WikiText-103, tokenized by each model's own
tokenizer.

\begin{table}[h]
\centering
\caption{Per-checkpoint protocol. Tied: whether input and output embeddings are
shared. Dtype: the stored precision (all tables analyzed in float32).
Tail: the norm of the shortest percentile of rows as a fraction of the
median norm, and the number of rows below half the median. $\rho_S$: the
Spearman correlation between row norm and log WikiText-103 count. Cov.:
the fraction of the vocabulary the corpus sees at least once. }
\label{tab:protocol}
\footnotesize
\begin{tabular}{lllllrrrr}
\toprule
Model &  Embeddings & Dtype & Multimodal & Tail & $<\tfrac12$ & $\rho_S$ & Cov. \\
\midrule
GPT-2            & tied & fp32 & no & $75\%$ & $0$ & $-0.46$ & $88\%$ \\
Pythia-160M      & untied & float16 & no & $84\%$ & $281$ & $+0.38$ & $68\%$ \\
Pythia-410M      & untied & float16 & no & $86\%$ & $278$ & $+0.44$ & $68\%$ \\
Pythia-1.4B      & untied & float16 & no & $85\%$ & $279$ & $+0.50$ & $68\%$ \\
Pythia-2.8B      & untied & float16 & no & $84\%$ & $278$ & $+0.50$ & $68\%$ \\
Pythia-6.9B      & untied & float16 & no & $79\%$ & $379$ & $+0.50$ & $68\%$ \\
Pythia-12B       & untied & float16 & no & $46\%$ & $634$ & $+0.51$ & $68\%$ \\
Qwen2.5-0.5B    & tied & bfloat16 & no & $65\%$ & $0$ & $-0.08$ & $34\%$ \\
Gemma-4         & tied & bfloat16 & yes & $100\%$ & $0$ & $-0.01$ & $33\%$ \\
K3              & untied & bfloat16 & yes & $55\%$ & $1208$ & $+0.50$ & $34\%$ \\
DeepSeek-V4-Pro  & untied & bfloat16 & no & $51\%$ & $1235$ & $-0.29$ & $44\%$ \\
Qwen3.8-27B      & untied & fp32 & yes & $39\%$ & $3924$ & $+0.39$ & $25\%$ \\
GLM-4.7         & untied & bfloat16 & no & $48\%$ & $1599$ & $+0.15$ & $36\%$ \\
\bottomrule
\end{tabular}
\end{table}

\begin{table}[h]
\centering
\caption{Checkpoint audit. $V$ and $D$ are the embedding matrix's rows and
width. ``Padded'' counts rows beyond the tokenizer's vocabulary. ``Zero'',
``tiny'', and ``dup.'' count zero-norm rows (norm below $10^{-8}$), rows with
norm below $1\%$ of the table's median row norm (zero-norm rows are a subset),
and exact-duplicate rows.}
\label{tab:checkpoints}
\footnotesize
\setlength{\tabcolsep}{3pt}
\begin{tabular}{llrrrrrr}
\toprule
Model & Checkpoint & $V$ & $D$ & Padded & Zero & Tiny & Dup. \\
\midrule
GPT-2           & \texttt{gpt2}                        & $50{,}257$  & $768$  & $0$   & $0$   & $0$   & $0$ \\
Pythia-160M     & \texttt{EleutherAI/pythia-160m}      & $50{,}304$  & $768$  & $27$  & $0$   & $238$ & $0$ \\
Pythia-410M     & \texttt{EleutherAI/pythia-410m}      & $50{,}304$  & $1024$ & $27$  & $0$   & $0$   & $0$ \\
Pythia-1.4B     & \texttt{EleutherAI/pythia-1.4b}      & $50{,}304$  & $2048$ & $27$  & $0$   & $0$   & $0$ \\
Pythia-2.8B     & \texttt{EleutherAI/pythia-2.8b}      & $50{,}304$  & $2560$ & $27$  & $0$   & $0$   & $0$ \\
Qwen2.5-0.5B    & \texttt{Qwen/Qwen2.5-0.5B}           & $151{,}936$ & $896$  & $271$ & $0$   & $0$   & $2$ \\
Gemma-4         & \texttt{google/gemma-4-31B}          & $262{,}144$ & $5376$ & $0$   & $0$   & $0$   & $0$ \\
K3              & \texttt{moonshotai/Kimi-K3}          & $163{,}840$ & $7168$ & $0$   & $0$   & $21$  & $0$ \\
DeepSeek-V4-Pro & \texttt{deepseek-ai/DeepSeek-V4-Pro} & $129{,}280$ & $7168$ & $0$   & $0$   & $0$   & $0$ \\
Qwen3.8-27B     & \texttt{Qwen/Qwen3.8-27B}            & $248{,}320$ & $5120$ & $243$ & $0$   & $1$   & $0$ \\
GLM-4.7         & \texttt{zai-org/GLM-4.7}             & $151{,}552$ & $5120$ & $187$ & $128$ & $485$ & $127$ \\
\bottomrule
\end{tabular}
\end{table}

\section{Supplementary figures}
\label{app:figs}

The calibration behind Figure~\ref{fig:sandbox_synth}: we planted a
$d$-dimensional bulk in $768$ ambient dimensions, added a hub clump with
the geometry measured in the real tables, and re-ran the instrument. With
no hub it reads $5/14/28/45$ at $d = 5/15/40/80$: accurate at low
dimension and increasingly low above $d \approx 15$, the known compression
of neighbor-ratio estimators, so cell readings above that are readings
rather than dimensions. With up to
$20\%$ hub rows the cell median stays at the planted value.

\begin{figure}[h]
\centering
\includegraphics[width=0.6\linewidth]{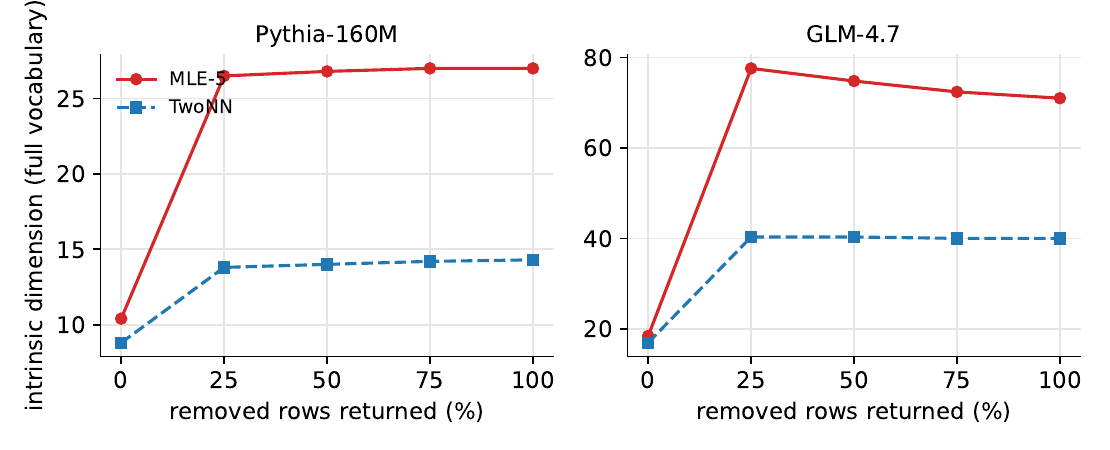}
\caption{The re-insertion sweep on Pythia-160M and GLM-4.7 (Table~\ref{tab:reinsert}).}
\label{fig:sandbox}
\end{figure}

\begin{figure}[h]
\centering
\includegraphics[width=\linewidth]{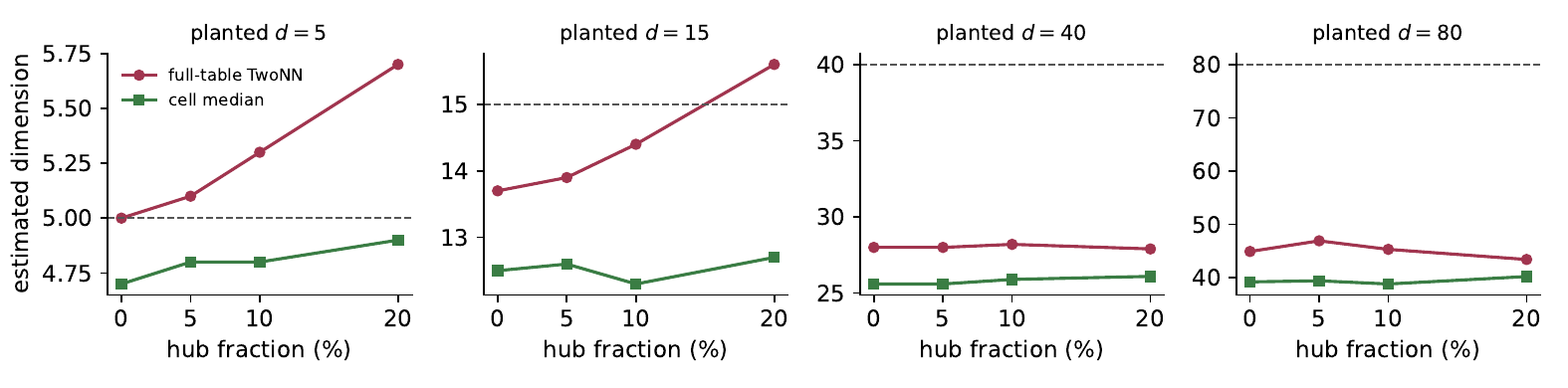}
\caption{Synthetic calibration. Each panel plants a $d$-dimensional bulk
in $D = 768$ with a hub clump of the measured geometry at the given hub
fraction. Dashed line: the planted dimension. The real-data counterpart
is the re-insertion sweep of Figure~\ref{fig:sandbox}.}
\label{fig:sandbox_synth}
\end{figure}

\end{document}